\documentclass[conference]{IEEEtran}
\IEEEoverridecommandlockouts
\usepackage{cite}
\usepackage{stfloats}
\usepackage{xcolor}
\usepackage{multirow}
\usepackage{stfloats}
\usepackage{amsmath,amssymb,amsfonts}
\usepackage{algorithmic}
\usepackage{graphicx}
\usepackage{textcomp}
\usepackage{xcolor}
\def\BibTeX{{\rm B\kern-.05em{\sc i\kern-.025em b}\kern-.08em
    T\kern-.1667em\lower.7ex\hbox{E}\kern-.125emX}}
\begin{document}

\title{Hybrid Kolmogorov-Arnold Network and XGBoost Framework for Week-Ahead Price Forecasting in Australia's National Electricity Market
}

\author{\IEEEauthorblockN{Houxuan Zhou\textsuperscript{1}, Sriram Prasad\textsuperscript{1}, Chenghao Huang\textsuperscript{1}, Jiajie Feng\textsuperscript{2}, Hao Wang\textsuperscript{1,3*}}
\IEEEauthorblockA{\textsuperscript{1}Department of Data Science and AI, Faculty of IT, Monash University, Australia \\
\textsuperscript{2}School of Electrical Engineering and Computer Science, University of Queensland, Australia\\
\textsuperscript{3}Monash Energy Institute, Monash University, Australia\\
}

\thanks{*Corresponding author: Hao Wang (hao.wang2@monash.edu).}
\thanks{This work was supported in part by the Australian Research Council (ARC) Discovery Early Career Researcher Award (DECRA) under Grant DE230100046.}
}

\maketitle

\begin{abstract}
Accurate electricity price forecasting (EPF) is essential for market participants to support operational planning and risk management, yet remains challenging due to strong volatility, nonlinear dynamics, and frequent extreme price spikes. These challenges are particularly pronounced in the Australian National Electricity Market (NEM), where high renewable penetration further increases uncertainty.
This paper investigates week-ahead electricity price forecasting and proposes a hybrid KAN+XGBoost framework that integrates Kolmogorov–Arnold Networks (KAN) with tree-based learning. 
The proposed approach combines the global nonlinear representation capability of KAN with the local robustness of XGBoost to capture both long-term dependencies and short-term price fluctuations.
Experiments are conducted on real-world NEM data using an expanding window evaluation strategy. The results demonstrate that the proposed model outperforms benchmark methods, including SARIMAX, Long Short-Term Memory (LSTM), standalone KAN, and XGBoost, reducing MAE by approximately 12\% compared to XGBoost and by over 50\% compared to a naive baseline. The results suggest that hybrid learning strategies provide an effective and robust solution for electricity price forecasting in highly dynamic electricity markets.
\end{abstract}

\begin{IEEEkeywords}
Electricity price forecasting, Kolmogorov-Arnold networks, XGBoost, hybrid models, ensemble learning, Australian National Electricity Market
\end{IEEEkeywords}

\section{Introduction}
Electricity price forecasting (EPF) plays a critical role in modern industrial energy systems, where accurate price signals are essential for market participation, risk management, and operational planning \cite{bunn2002forecasting}. Industrial stakeholders, including generators and large industrial consumers, rely on reliable forecasting to support resource scheduling and financial decision-making. Electricity prices are inherently volatile and influenced by complex supply–demand dynamics, including demand fluctuations, weather conditions, renewable generations, generation costs, and network constraints \cite{hameer2015review}.
These challenges are particularly pronounced in Australia’s National Electricity Market (NEM), where prices are settled at five-minute resolution, leading to increased short-term volatility. Recent studies have also highlighted the growing importance of flexibility management and market coordination in modern electricity systems with high renewable penetration \cite{11264331}. In addition, the growing penetration of variable renewable energy (VRE), such as wind and solar, further amplifies this complexity.

VRE sources are dispatched first under the merit-order effect due to their near-zero marginal costs. However, their intermittent nature increases market uncertainty and contributes to nonlinear price dynamics and abrupt fluctuations under changing supply-demand conditions \cite{guerra2022facing,csereklyei2019effect}. As a result, EPF in the NEM requires models capable of capturing both complex nonlinear dependencies and high market volatility. Although day-ahead forecasting has been widely studied in the NEM, week-ahead hourly forecasting remains relatively underexplored despite its practical importance. Compared with short-term forecasting, week-ahead prediction provides longer lead time for bidding, generation scheduling, and risk hedging, making it particularly valuable under increasing renewable penetration and medium-term operational planning \cite{ziel2018probabilistic,weron2014electricity}.

Traditional statistical models, such as Autoregressive Integrated Moving Average (ARIMA) \cite{box2015time} and Generalized Autoregressive Conditional Heteroskedasticity (GARCH) \cite{bollerslev1986generalized}, can capture seasonality and volatility patterns, but they often struggle with nonlinear dynamics and extreme price spikes  \cite{misiorek2006point,weron2014electricity}. Extensions, such as Seasonal Autoregressive Integrated Moving Average with Exogenous Variables (SARIMAX) and Vector AutoRegression (VAR) models, further incorporate exogenous variables and interdependencies, improving modeling flexibility in electricity price forecasting \cite{box2015time, sims1980macroeconomics,weron2014electricity}. 

More recently, machine learning and deep learning approaches have gained increasing attention. 
Neural network architectures, including Long Short-Term Memory (LSTM) \cite{hochreiter1997long} and Gated Recurrent Unit (GRU) \cite{cho2014learning}, have shown a strong capability in modeling nonlinear relationships and temporal dependencies, often outperforming traditional statistical models
\cite{lago2018forecasting,lago2021forecasting,fang2021survey}. In particular, recurrent models are well-suited for capturing sequential patterns and long-term dependencies in time-series data. However, their performance remains sensitive to noise and instability when applied to longer forecasting horizons and highly volatile market conditions \cite{lago2018forecasting,lago2021forecasting}. 

Furthermore, hybrid and ensemble approaches have been explored to improve forecasting performance by leveraging complementary model strengths \cite{maciejowska2020pca,neupane2017ensemble}. While these approaches have shown promising results, they are often designed for short-term forecasting settings and may not fully address the challenges of capturing both global structures and local irregular variations in more complex environments such as the NEM.
Consequently, existing approaches typically face a trade-off between global representation and local robustness, making them insufficient for reliable week-ahead forecasting in the NEM. This limitation highlights the need for more effective models that can capture both global and local price dynamics.

Kolmogorov-Arnold Networks (KAN), based on the Kolmogorov-Arnold representation theorem \cite{liu2024kan}, provides a flexible functional approximation framework by learning univariate transformations. This design enables KAN to effectively capture complex nonlinear dependencies in electricity price series, making it particularly suitable for modeling global functional relationships in EPF. In addition, its use of univariate functions improves interpretability compared to conventional neural networks, and recent studies have demonstrated its competitive performance in time-series forecasting tasks \cite{vaca2024kolmogorov}.
In contrast, tree-based models such as XGBoost are inherently robust to local irregularities and outliers due to their piecewise partitioning mechanism, which allows them to effectively handle abrupt changes in data \cite{chen2016xgboost}. However, their piecewise nature may limit their ability to capture smooth global dependencies and long-term temporal patterns over extended forecasting horizons.
The complementary characteristics of KAN and XGBoost highlight the limitations of using either model independently. KAN may struggle with localized irregularities and extreme price spikes, while XGBoost is less effective in capturing smooth global dependencies over longer horizons. Motivated by this challenge, this paper proposes a hybrid KAN+XGBoost framework for week-ahead electricity price forecasting in the NEM. From a functional decomposition perspective, KAN is used for learning global nonlinear structures, while XGBoost is effective in modeling residual local deviations, thus enabling a balanced representation of long-term dependencies and short-term irregular variations.
\begin{itemize}
\item This study addresses the challenge of week-ahead electricity price forecasting in the NEM, where models must simultaneously handle long-term dependencies and short-term extreme price variations. The proposed framework provides longer-term foresight that can support more informed decision-making for market participants.

\item A hybrid KAN+XGBoost framework is proposed, which integrates functional representation learning with tree-based partitioning. This design enables the model to jointly capture global nonlinear structures and local irregular variations in electricity price dynamics.

\item Extensive experiments on real-world NEM data demonstrate that the proposed approach consistently outperforms statistical, machine learning, and deep learning baselines across multiple states. The model achieves approximately 10-15\% improvement in MAE over strong benchmarks, while maintaining robust performance under diverse market conditions.
\end{itemize}

The remainder of this paper is organized as follows. Section~\ref{Method} introduces the methodology. Section~\ref{Exp} describes the experimental setup. Section~\ref{Result} presents and discusses the results, and Section~\ref{Conclu} concludes the paper.

\section{KAN–XGBoost Forecasting Framework}\label{Method}
\begin{figure}[!t]
\centering
\includegraphics[width=0.8\columnwidth]{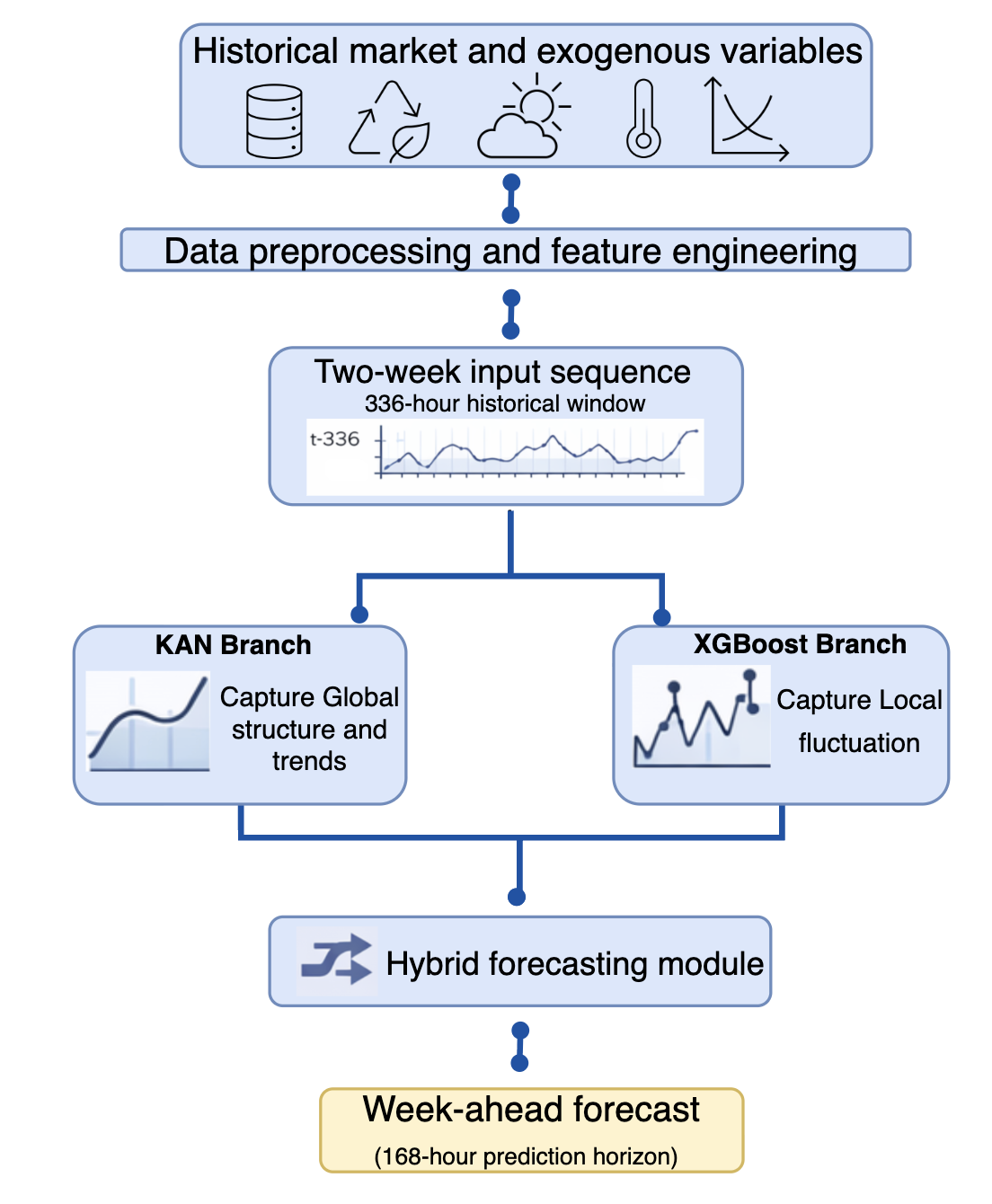}
\caption{Overview of the proposed KAN+XGBoost hybrid forecasting framework.}
\label{fig:framework}
\end{figure}

This section presents the proposed framework for week-ahead electricity price forecasting in the NEM. We first define the forecasting problem, followed by data preparation and feature engineering procedures. Finally, the proposed hybrid KAN–XGBoost model is introduced. An overview of the proposed framework is illustrated in Fig.~\ref{fig:framework}.

\subsection{Problem Formulation}

Let $y_t$ denote the electricity price at time $t$, and let $\mathbf{z}_t$ represent the corresponding exogenous variables, including demand, renewable generation, and weather features.

Given a historical input window of length $L$, the model input is defined as:
\begin{equation}
\mathbf{x}_t = \{\mathbf{y}_{t-L+1:t}, \mathbf{z}_{t-L+1:t}\},
\end{equation}
where $\mathbf{y}_{t-L+1:t}$ denotes the historical electricity price sequence and $\mathbf{z}_{t-L+1:t}$ represents the corresponding exogenous variables.

The forecasting task is to learn a mapping function:
\begin{equation}
\hat{\mathbf{y}}_{t+1:t+H} = g(\mathbf{x}_t),
\end{equation}
where $L=336$ represents a two-week look-back window and $H=168$ corresponds to the week-ahead forecasting horizon. The objective is to accurately predict the future electricity price trajectory over the next week.

\subsection{Data and Feature Preparation}

Given the complexity of modeling electricity prices in the NEM, multiple variables and dependencies should be captured to effectively develop week-ahead EPF models.

Feature selection is guided by domain knowledge of electricity market dynamics, with variables chosen to reflect both temporal dependencies and supply–demand interactions in the NEM. 

The model incorporates lagged electricity prices to capture strong autocorrelation and recurring daily and weekly patterns, with a 168-hour lag representing weekly seasonality. Total demand is included as a key driver of short-term price fluctuations due to demand inelasticity. 

Supply-side variables, including net interchange and renewable generation forecasts (solar and wind), are incorporated to capture cross-regional electricity flows and the impact of variable renewable energy on price formation. Temperature is included as a proxy for demand variation driven by weather conditions.

To stabilize variance and mitigate the influence of extreme price spikes, an inverse hyperbolic sine transformation is applied to the target variable.

\subsection{Model Framework}
\subsubsection{KAN}

KAN is a neural architecture based on the Kolmogorov-Arnold representation theorem \cite{liu2024kan}, which states that a multivariate function can be expressed as a composition of univariate functions:
\begin{equation}
f(\mathbf{x}) = \sum_{q=0}^{2n} \Phi_q \left( \sum_{p=1}^{n} \phi_{q,p}(x_p) \right),
\end{equation}
where $n$ denotes the input dimensionality. KAN uses spline-based learnable activation functions enabling adaptive basis representation.

In this study, KAN learns a nonlinear mapping function $g_{\text{KAN}}(\cdot)$ to approximate the forecasting relationship:
\begin{equation}
\hat{\mathbf{y}}_{t+1:t+H}^{\text{KAN}} = g_{\text{KAN}}(\mathbf{x}_t).
\end{equation}

The implemented KAN model uses two hidden layers, with spline transformations parameterized by a grid size of 3 and spline order of 3. 
A min-max scaler is applied for feature normalization. 
The model is trained using the MAE loss function for up to 300 optimization steps, with early stopping applied if validation performance does not improve for 30 steps. 
All forecasts are generated at an hourly frequency.

KAN is primarily responsible for learning the global nonlinear structure of electricity price dynamics, including smooth temporal patterns and interactions between historical prices and exogenous drivers.

\subsubsection{XGBoost}
XGBoost is a gradient boosting framework based on decision trees, known for its robustness to noise and strong performance on structured data. 
It builds an ensemble of decision trees sequentially, where each tree is trained to correct the residual errors of the previous ones. 

In this study, XGBoost uses the same input features as KAN and is trained to capture local patterns and short-term irregularities in electricity prices.  
Similarly, XGBoost is employed to approximate the forecasting function and provides a complementary prediction:
\begin{equation}
\hat{\mathbf{y}}_{t+1:t+H}^{\text{XGB}} = g_{\text{XGB}}(\mathbf{x}_t).
\end{equation}
Its tree-based structure enables effective modeling of abrupt changes and nonlinear interactions that may not be fully captured by neural networks.

\subsubsection{Hybrid KAN+XGBoost Model}

The hybrid formulation leverages the complementary strengths of KAN and XGBoost, where KAN captures global nonlinear structures while XGBoost models local variations and abrupt changes. The final prediction is obtained through a weighted combination:
\begin{equation}
\hat{\mathbf{y}}_{t+1:t+H} = \alpha \hat{\mathbf{y}}_{t+1:t+H}^{\text{KAN}} + (1 - \alpha)\hat{\mathbf{y}}_{t+1:t+H}^{\text{XGB}}.
\end{equation}

The weighting coefficient $\alpha \in [0,1]$ is selected to minimize the validation error:
\begin{equation}
\alpha^* = \arg\min_{\alpha \in [0,1]} \mathcal{L}\left( \alpha \hat{\mathbf{y}}^{\text{KAN}} + (1-\alpha)\hat{\mathbf{y}}^{\text{XGB}}, \mathbf{y} \right).
\end{equation}

The ensemble weights are selected based on validation performance using multiple weight combinations. The 0.5/0.5 setting achieves the best overall validation performance and is therefore adopted across all NEM states.
This formulation enables the model to capture both global nonlinear structures and local irregular variations, improving robustness under volatile market conditions.

\section{Experiment}\label{Exp}

\subsection{Dataset Description}

The dataset is obtained from the Australian Energy Market Operator (AEMO) \cite{aemo} and World Weather Online \cite{weather}, covering the period from April 2024 to March 2025 with diverse seasonal and volatility regimes. 
It includes electricity price (RRP), total demand, net interchange, renewable generation forecasts (solar and wind), and weather-related variables such as temperature, wind speed, humidity, and cloud cover. 

Weather variables play an important role in electricity price formation by influencing both demand patterns and renewable generation output. In particular, temperature affects electricity consumption, while wind and solar conditions directly impact renewable generation, thereby contributing to price variability.

Raw AEMO data are provided at a five-minute resolution and are aggregated to hourly values by averaging. 
All variables are aligned to an hourly temporal resolution to ensure consistency across inputs.

The study considers the five regions of the National Electricity Market (NEM), namely NSW, VIC, QLD, SA, and TAS. Weather data are collected from capital cities to represent major demand centers.

No outlier removal is applied, as extreme price spikes are inherent characteristics of the NEM and are essential for evaluating model robustness.
\subsection{Experimental Setup}
An expanding window evaluation strategy is adopted, with four consecutive weekly test sets from March 1 to March 28, 2025.
For each test week, models are trained using all historical data available from April 2024 up to the forecast period, ensuring a strict temporal separation between training and evaluation.
This process is repeated across multiple test weeks to assess model robustness under varying market conditions.

The forecasting task is to predict electricity prices for the next 168 hours (one week) using a 336-hour (two-week) input window. 
Although the original AEMO data are provided at a five-minute resolution, all variables are aggregated to hourly values for modeling. 
Accordingly, the forecasting horizon is defined at an hourly resolution.

The models generate direct multi-step forecasts for the entire horizon, avoiding recursive error accumulation and enabling stable long-horizon prediction.

The input features include lagged electricity prices (up to 168 hours), total demand, net interchange, solar and wind generation forecasts, and temperature variables, selected based on their influence on electricity price formation.
All input variables are normalized using min-max scaling.

For the KAN model, we adopt a two-layer architecture with a grid size of 3 and spline order of 3. The model is trained using the Adam optimizer with a learning rate of 0.001. Early stopping is applied with a patience of 30 steps to prevent overfitting.

For the XGBoost model, we use the squared error objective with 500 estimators, a maximum tree depth of 6, and a learning rate of 0.05. Sub-sampling and column sampling rates are set to 0.8 to improve generalization. L2 regularization is applied to control model complexity.

All models are trained under the same experimental settings to ensure a fair comparison.

\subsection{Baselines}
The proposed hybrid model is compared against individual KAN and XGBoost models to evaluate the contribution of each component. In addition, a naïve forecasting baseline, a statistical model (SARIMAX), and a deep learning model (LSTM) are included for broader comparison.

All models are trained using the same input features and evaluation settings to ensure a fair comparison. The naïve model assumes that future prices are equal to the most recent observed value. The SARIMAX model incorporates exogenous variables consistent with the proposed framework, while the LSTM model is implemented in a sequence-to-sequence manner to produce direct multi-step forecasts.

\subsection{Evaluation Metrics}
Model performance is evaluated using Mean Absolute Error (MAE) and Root Mean Squared Error (RMSE). 
In addition, relative MAE (rMAE) is reported against a naïve baseline to assess improvement over simple forecasting. 
MAE measures the average magnitude of errors, while RMSE places greater emphasis on large deviations, which is important in volatile electricity price series.

\section{Results and Discussion}\label{Result}
The overall forecasting performance is presented in Table~\ref{tabel:overall}. 
KAN+XGBoost achieves the lowest MAE, RMSE, and rMAE among all compared methods, indicating improved overall accuracy and robustness. 
Specifically, the proposed model reduces MAE by approximately 12.0\% compared to XGBoost and more than 50\% relative to the naive benchmark, demonstrating the effectiveness of the learning-based approach.

\begin{table}[t]
\centering
\caption{Overall forecasting performance comparison}
\renewcommand{\arraystretch}{1.3} 
\setlength{\tabcolsep}{10pt} 
\label{tabel:overall}
\begin{tabular*}{\columnwidth}{@{\extracolsep{\fill}}lccc}
\hline
\textbf{Model} & \textbf{MAE} & \textbf{RMSE} & \textbf{rMAE} \\
\hline
\textbf{KAN+XGBoost} & \textbf{26.17} & \textbf{81.90} & \textbf{0.498} \\
XGBoost         & 29.49 & 90.05 & 0.561 \\
LSTM            & 30.90 & 86.90& 0.588 \\
KAN             & 33.49 & 89.61 & 0.638 \\
SARIMAX         & 36.42 & 90.92 & 0.694 \\
Naive           & 52.50 & 132.58 & 1.000 \\
\hline
\end{tabular*}
\end{table}

Across individual NEM states (Table~\ref{tab:state_performance}), KAN+XGBoost consistently maintains strong performance, highlighting its ability to generalize under varying regional market conditions. 
This is particularly evident in relatively stable regions such as QLD and VIC, where substantial error reductions are observed. 
In highly volatile regions such as NSW, the model remains effective under frequent price spikes, achieving the lowest MAE and RMSE among all methods. 
In SA, XGBoost achieves slightly lower RMSE but higher MAE than KAN+XGBoost, suggesting higher sensitivity to extreme values but reduced overall prediction stability.

\begin{table}[t]
\centering
\caption{State-wise forecasting performance across five NEM states (best values in bold)}
\label{tab:state_performance}

\scriptsize  
\setlength{\tabcolsep}{3pt}  
\renewcommand{\arraystretch}{1.2}

\begin{tabular}{lcccccc}
\hline
\textbf{State} & \textbf{Metric} & \textbf{KAN+XGBoost} & \textbf{XGBoost} & \textbf{KAN} & \textbf{SARIMAX} & \textbf{LSTM} \\
\hline

\multirow{3}{*}{NSW}
& MAE  & \textbf{35.53} & 40.38 & 37.52 & 62.30 & 37.29 \\
& RMSE & \textbf{165.7} & 175.2 & 175.8 & 182.3 & 171.9 \\
& rMAE & \textbf{0.567} & 0.644 & 0.599 & 0.99 & 0.595 \\
\hline

\multirow{3}{*}{QLD}
& MAE  & \textbf{22.57} & 28.23 & 27.96 & 34.36 & 25.02 \\
& RMSE & \textbf{52.50} & 79.20 & 58.04 & 61.96 & 60.16 \\
& rMAE & \textbf{0.630} & 0.789 & 0.781 & 0.966 & 0.698 \\
\hline

\multirow{3}{*}{SA}
& MAE  & \textbf{26.22} & 28.50 & 39.73 & 29.70 & 33.14 \\
& RMSE & 37.19 & \textbf{36.88} & 50.81 & 40.38 & 42.09 \\
& rMAE & \textbf{0.356} & 0.387 & 0.540 & 0.403 & 0.450 \\
\hline

\multirow{3}{*}{TAS}
& MAE  & \textbf{23.78} & 25.94 & 33.60 & 24.72 & 31.59 \\
& RMSE & \textbf{32.25} & 33.97 & 44.51 & 33.80 & 40.87 \\
& rMAE & \textbf{0.660} & 0.706 & 0.915 & 0.672 & 0.860 \\
\hline

\multirow{3}{*}{VIC}
& MAE  & \textbf{22.73} & 24.42 & 28.67 & 31.03 & 27.49 \\
& RMSE & \textbf{29.26} & 33.18 & 36.20 & 38.66 & 34.52 \\
& rMAE & \textbf{0.423} & 0.455 & 0.534 & 0.578 & 0.512 \\
\hline

\end{tabular}
\end{table}

\begin{figure*}[t]

\centering
\includegraphics[width=0.7\linewidth]{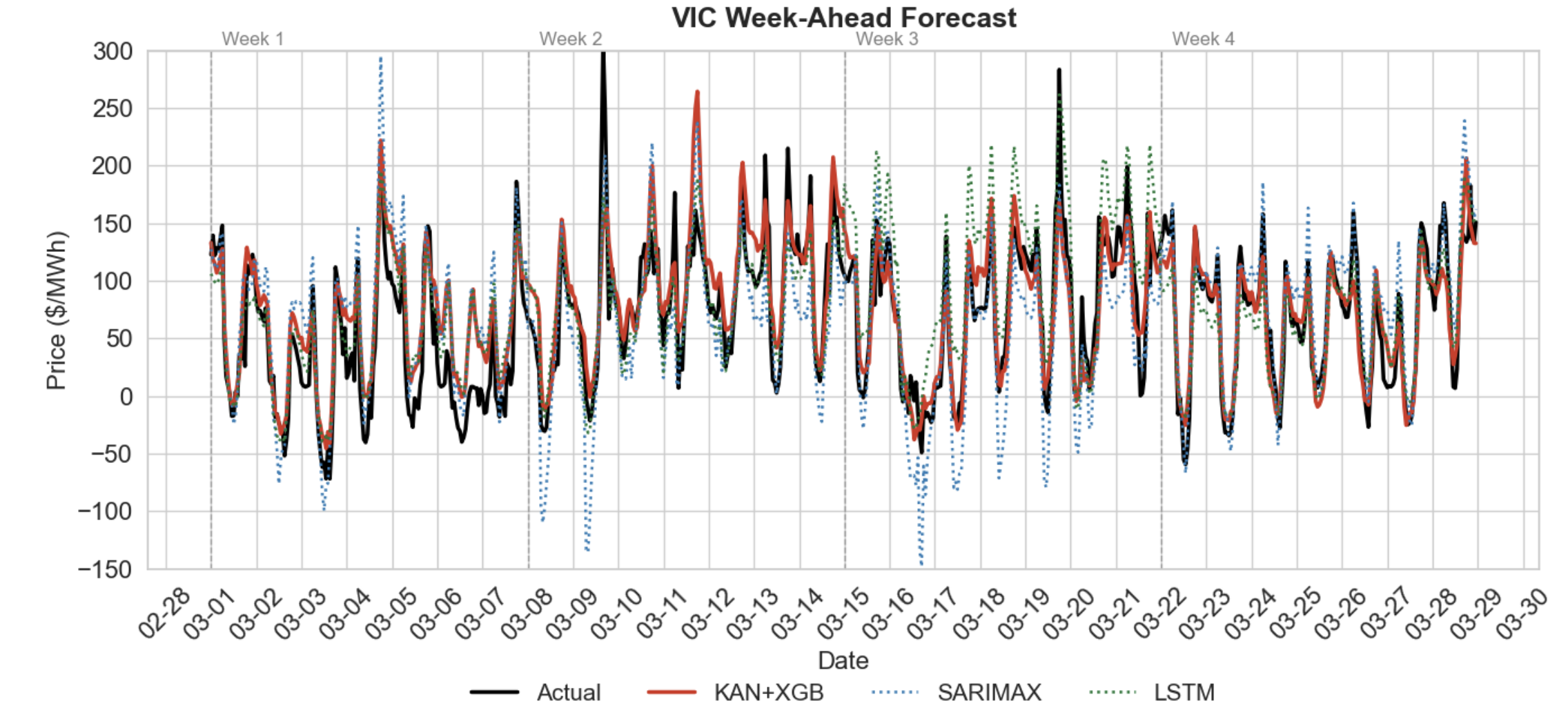}
\caption{Comparison of week-ahead electricity price forecasting performance in VIC over four evaluation weeks.}
\label{fig:VIC}
\end{figure*}

Figure~\ref{fig:VIC} illustrates the week-ahead forecasting results for VIC over a representative period. 
While all models capture the general trend, KAN+XGBoost demonstrates more stable tracking during fluctuations, indicating improved adaptability to short-term variations.
From a modeling perspective, the improved performance of KAN+XGBoost can be attributed to the complementary strengths of its components. 
KAN captures global nonlinear structures and long-term dependencies, whereas XGBoost is more robust to local irregularities and abrupt changes. 
Their integration enables a balanced representation of long-term trends and short-term volatility, which is critical for electricity price forecasting in highly dynamic markets.

To further validate the effectiveness of the proposed hybrid design, an ablation study with different ensemble weightings is conducted (Table~\ref{tab:ablation}). 
The ablation results show that the equal-weight configuration (0.5/0.5) achieves the best overall performance, indicating that a balanced integration of global nonlinear learning and local pattern refinement is most effective, while over-reliance on either component leads to performance degradation.

In addition, a KAN+LSTM hybrid model is evaluated for comparison. The results show that the proposed KAN+XGBoost framework achieved lower MAE and rMAE values overall, suggesting more effective complementary learning behavior for the current forecasting task.
\begin{table}[t]
\centering
\caption{Ablation study on different ensemble weight settings}
\label{tab:ablation}
\renewcommand{\arraystretch}{1.3} 
\setlength{\tabcolsep}{10pt} 
\begin{tabular*}{\columnwidth}{@{\extracolsep{\fill}}lccc}
\hline
\textbf{Model} & \textbf{MAE} & \textbf{RMSE} & \textbf{rMAE} \\
\hline

\textbf{KAN+XGBoost (0.5/0.5)} & \textbf{26.17} & \textbf{81.90} & \textbf{0.498} \\
KAN+XGBoost (0.2/0.8) & 27.50 & 87.23 & 0.524 \\
KAN+XGBoost (0.6/0.4) & 28.00 & 85.65 & 0.533 \\
KAN+XGBoost (0.7/0.3) & 28.93 & 86.13 & 0.551 \\
KAN+LSTM (0.5/0.5) & 29.28 & 86.54 & 0.558 \\
XGBoost & 29.49 & 90.05 & 0.558 \\
KAN & 33.49 & 89.61 & 0.638 \\

\hline
\end{tabular*}
\end{table}
The effectiveness of the proposed approach is further linked to the heterogeneous characteristics of the NEM. 
Different states exhibit distinct behaviors in terms of volatility, generation mix, and frequency of price spikes, making it difficult for a single model to perform consistently across all regions. 
The hybrid framework addresses this challenge by leveraging complementary learning mechanisms, enabling robust performance without requiring fully state-specific model redesign.

Despite the overall improvements, accurately capturing extreme price spikes remains a challenge for all models. 
During high-volatility periods, predictions tend to underestimate extreme values, which may be attributed to the use of loss functions such as MAE and MSE that prioritize average performance over rare extreme deviations.

To further evaluate performance under extreme market conditions, spike events are analyzed using an EVT/POT-based thresholding approach \cite{bystrom2005extreme}. 
In the main analysis, we focus on the 0.95 tail threshold, as it provides a balance between extremity and a sufficient number of exceedances for reliable evaluation. 

As shown in Table~\ref{tab:evt_main} the proposed KAN+XGBoost model achieves the lowest extreme-event MAE among all compared methods, indicating improved robustness in capturing high-price events.

\begin{table}[t]
\centering
\caption{Overall and EVT/POT-based extreme-event forecasting performance}
\label{tab:evt_main}

\resizebox{0.9\columnwidth}{!}{%
\begin{tabular}{lcc}
\hline
\textbf{Model} & \textbf{Overall MAE} & \textbf{Extreme MAE (EVT 0.95)} \\
\hline
KAN+XGBoost & \textbf{26.17} & \textbf{103.88} \\
XGBoost     & 29.49 & 111.10 \\
LSTM        & 30.90 & 109.22 \\
KAN         & 33.49 & 120.76 \\
SARIMAX     & 36.42 & 116.81 \\
Naive       & 52.50 & 188.81 \\
\hline
\end{tabular}%
}
\end{table}

Finally, the relatively weaker performance of KAN as a standalone model highlights its limitations in highly dynamic environments. 
While KAN is capable of modeling complex nonlinear relationships, its sensitivity to noise and training instability can lead to inconsistent predictions. 
However, these limitations are mitigated when combined with XGBoost, suggesting that KAN is more effective as part of a hybrid framework.
Overall, the results indicate that hybrid modeling strategies provide a robust and effective solution for electricity price forecasting in highly dynamic markets such as the NEM.

\subsection{Implications, Limitations and Future Work}
The findings of this study highlight the effectiveness of hybrid modeling strategies in highly volatile environments.
By combining complementary learning mechanisms, the proposed approach improves robustness and generalization, which is critical for real-world electricity market applications 

Despite the promising results, several limitations should be acknowledged.
The study assumes the availability of accurate exogenous variables such as demand and weather data. In practice, these inputs are forecasted and may introduce additional uncertainty.
In addition, the dataset only covers the period from April 2024 to March 2025, which may limit the model's ability to capture seasonal effects and rare market regimes.
Extreme price spikes also remain difficult to predict due to their rare and irregular nature.

Future work can extend this study by incorporating probabilistic forecasting or quantile-based approaches to better capture uncertainty and extreme events.
Adaptive ensemble strategies and the integration of richer domain-specific features, such as generator outages or bidding behavior, may further improve predictive performance.
Extending the framework to longer forecasting horizons also represents a promising direction for future research.

\section{Conclusion}\label{Conclu}

This study investigated week-ahead electricity price forecasting in the Australian National Electricity Market using statistical, deep learning, and hybrid models.
The proposed KAN+XGBoost framework achieves the best overall forecasting performance among all compared methods, reducing MAE from 29.49 to 26.17 compared to XGBoost.
The improved performance of KAN+XGBoost highlights the effectiveness of hybrid modeling strategies in capturing both long-term nonlinear relationships and short-term volatility in electricity prices. It also verifies that functional decomposition-based hybrid learning is a promising direction.
The proposed approach demonstrates strong generalization across different NEM states, suggesting that model complementarity can reduce the need for fully state-specific configurations.

Despite these improvements, accurately predicting extreme price spikes remains challenging.
Future work will focus on probabilistic forecasting, adaptive ensemble strategies, and incorporating richer domain-specific features to further improve performance.
Overall, the findings suggest that hybrid learning frameworks provide a robust and practical solution for electricity price forecasting in highly dynamic markets.

\bibliography{ref.bib} 

@article{bunn2002forecasting,
  title={Forecasting loads and prices in competitive power markets},
  author={Bunn, Derek W},
  journal={Proceedings of the IEEE},
  volume={88},
  number={2},
  pages={163--169},
  year={2002},
  publisher={IEEE}
}

@article{hameer2015review,
  title={A review of large-scale electrical energy storage},
  author={Hameer, Sameer and Van Niekerk, Johannes L},
  journal={International journal of energy research},
  volume={39},
  number={9},
  pages={1179--1195},
  year={2015},
  publisher={Wiley Online Library}
}

@article{csereklyei2019effect,
  title={The effect of wind and solar power generation on wholesale electricity prices in Australia},
  author={Csereklyei, Zsuzsanna and Qu, Songze and Ancev, Tihomir},
  journal={Energy Policy},
  volume={131},
  pages={358--369},
  year={2019},
  publisher={Elsevier}
}

@article{guerra2022facing,
  title={Facing the high share of variable renewable energy in the power system: Flexibility and stability requirements},
  author={Guerra, K and Haro, P and Guti{\'e}rrez, RE and G{\'o}mez-Barea, A},
  journal={Applied Energy},
  volume={310},
  pages={118561},
  year={2022},
  publisher={Elsevier}
}

@article{liu2024kan,
  title={Kan: Kolmogorov-arnold networks},
  author={Liu, Ziming and Wang, Yixuan and Vaidya, Sachin and Ruehle, Fabian and Halverson, James and Solja{\v{c}}i{\'c}, Marin and Hou, Thomas Y and Tegmark, Max},
  journal={arXiv preprint arXiv:2404.19756},
  year={2024}
}

@article{misiorek2006point,
  title={Point and interval forecasting of spot electricity prices: Linear vs. non-linear time series models},
  author={Misiorek, Adam and Trueck, Stefan and Weron, Rafal},
  journal={Studies in nonlinear dynamics and econometrics},
  volume={10},
  number={3},
  pages={0--34},
  year={2006},
  publisher={Walter de Gruyter}
}

@article{lago2021forecasting,
  title={Forecasting day-ahead electricity prices: A review of state-of-the-art algorithms, best practices and an open-access benchmark},
  author={Lago, Jesus and Marcjasz, Grzegorz and De Schutter, Bart and Weron, Rafa{\l}},
  journal={Applied Energy},
  volume={293},
  pages={116983},
  year={2021},
  publisher={Elsevier}
}

@article{fang2021survey,
  title={Survey on research of RNN-based spatio-temporal sequence prediction algorithms},
  author={Fang, Wei and Chen, Yupeng and Xue, Qiongying},
  journal={Journal on Big Data},
  volume={3},
  number={3},
  pages={97},
  year={2021},
  publisher={Tech Science Press}
}

@article{lago2018forecasting,
  title={Forecasting spot electricity prices: Deep learning approaches and empirical comparison of traditional algorithms},
  author={Lago, Jesus and De Ridder, Fjo and De Schutter, Bart},
  journal={Applied Energy},
  volume={221},
  pages={386--405},
  year={2018},
  publisher={Elsevier}
}

@article{maciejowska2020pca,
  title={PCA forecast averaging—predicting day-ahead and intraday electricity prices},
  author={Maciejowska, Katarzyna and Uniejewski, Bartosz and Serafin, Tomasz},
  journal={Energies},
  volume={13},
  number={14},
  pages={3530},
  year={2020},
  publisher={MDPI}
}

@article{neupane2017ensemble,
  title={Ensemble prediction model with expert selection for electricity price forecasting},
  author={Neupane, Bijay and Woon, Wei Lee and Aung, Zeyar},
  journal={Energies},
  volume={10},
  number={1},
  pages={77},
  year={2017},
  publisher={MDPI}
}

@article{weron2014electricity,
  title={Electricity price forecasting: A review of the state-of-the-art with a look into the future},
  author={Weron, Rafa{\l}},
  journal={International journal of forecasting},
  volume={30},
  number={4},
  pages={1030--1081},
  year={2014},
  publisher={Elsevier}
}

@inproceedings{vaca2024kolmogorov,
  title={Kolmogorov-arnold networks (kans) for time series analysis},
  author={Vaca-Rubio, Cristian J and Blanco, Luis and Pereira, Roberto and Caus, M{\`a}rius},
  booktitle={2024 IEEE Globecom Workshops (GC Wkshps)},
  pages={1--6},
  year={2024},
  organization={IEEE}
}

@misc{weather,
  author       = {{World Weather Online}},
  title        = {Weather API},
  howpublished = {\url{https://www.worldweatheronline.com/weather-api/}},
  note         = {Accessed: 2026-04-09}
}

@misc{aemo,
  author       = {{Australian Energy Market Operator (AEMO)}},
  title        = {Aggregated Data - National Electricity Market},
  howpublished = {\url{https://www.aemo.com.au/energy-systems/electricity/national-electricity-market-nem/data-nem/aggregated-data}},
  note         = {Accessed: 2026-04-09}
}

@article{ziel2018probabilistic,
  title={Probabilistic mid-and long-term electricity price forecasting},
  author={Ziel, Florian and Steinert, Rick},
  journal={Renewable and Sustainable Energy Reviews},
  volume={94},
  pages={251--266},
  year={2018},
  publisher={Elsevier}
}

@article{bystrom2005extreme,
  title={Extreme value theory and extremely large electricity price changes},
  author={Bystr{\"o}m, Hans NE},
  journal={International Review of Economics \& Finance},
  volume={14},
  number={1},
  pages={41--55},
  year={2005},
  publisher={Elsevier}
}

@inproceedings{chen2016xgboost,
  title={Xgboost: A scalable tree boosting system},
  author={Chen, Tianqi and Guestrin, Carlos},
  booktitle={Proceedings of the 22nd acm sigkdd international conference on knowledge discovery and data mining},
  pages={785--794},
  year={2016}
}

@book{box2015time,
  title={Time series analysis: forecasting and control},
  author={Box, George EP and Jenkins, Gwilym M and Reinsel, Gregory C and Ljung, Greta M},
  year={2015},
  publisher={John Wiley \& Sons}
}

@article{hochreiter1997long,
  title={Long short-term memory},
  author={Hochreiter, Sepp and Schmidhuber, J{\"u}rgen},
  journal={Neural computation},
  volume={9},
  number={8},
  pages={1735--1780},
  year={1997},
  publisher={MIT press}
}

@ARTICLE{11264331,
  author={Nikkhah, Mohammad Hassan and Azim, M. Imran and Amani, Ali Moradi and Jalili, Mahdi and Yu, Xinghuo},
  journal={IEEE Transactions on Industrial Informatics}, 
  title={Joint Upstream-Distribution Flexibility Mechanism Using Distributed Energy Storage Systems}, 
  year={2026},
  volume={22},
  number={2},
  pages={1269-1280},
  doi={10.1109/TII.2025.3625184}}

@article{bollerslev1986generalized,
  title={Generalized autoregressive conditional heteroskedasticity},
  author={Bollerslev, Tim},
  journal={Journal of econometrics},
  volume={31},
  number={3},
  pages={307--327},
  year={1986},
  publisher={Elsevier}
}

@article{sims1980macroeconomics,
  title={Macroeconomics and reality},
  author={Sims, Christopher A},
  journal={Econometrica: journal of the Econometric Society},
  pages={1--48},
  year={1980},
  publisher={JSTOR}
}

@inproceedings{cho2014learning,
  title={Learning phrase representations using RNN encoder--decoder for statistical machine translation},
  author={Cho, Kyunghyun and Van Merri{\"e}nboer, Bart and Gul{\c{c}}ehre, {\c{C}}a{\u{g}}lar and Bahdanau, Dzmitry and Bougares, Fethi and Schwenk, Holger and Bengio, Yoshua},
  booktitle={Proceedings of the 2014 conference on empirical methods in natural language processing (EMNLP)},
  pages={1724--1734},
  year={2014}
}

\end{document}